\documentclass[letterpaper]{article}

\usepackage[margin=1in]{geometry}

\newcommand{\rev}[1]{#1}

\usepackage{times}
\usepackage{helvet}
\usepackage{courier}
\usepackage[hyphens]{url}
\usepackage{graphicx}
\graphicspath{{Images/}}
\usepackage{natbib}
\usepackage{caption}
\usepackage{newfloat}
\usepackage{listings}
\DeclareCaptionStyle{ruled}{labelfont=normalfont,labelsep=colon,strut=off}
\usepackage{latexsym}
\usepackage{amssymb}
\usepackage{amsmath}
\usepackage{amsthm}
\usepackage{booktabs}
\usepackage{enumitem}
\usepackage{color}
\usepackage{hyperref}
\usepackage{subcaption}

\usepackage{algorithm,algpseudocode}
\algnewcommand{\algorithmicforeach}{\textbf{for each}}
\algdef{SE}[FOR]{ForEach}{EndForEach}[1]
  {\algorithmicforeach\ #1\ \algorithmicdo}
  {\algorithmicend\ \algorithmicforeach}

\title{FlowRL: Flow-Augmented Few-Shot Reinforcement Learning for Semi-Structured Sensor Data}
\author{
    Mohammad Pivezhandi$^1$ \and Abusayeed Saifullah$^2$ \\[1em]
    $^1$Wayne State University, Detroit, MI, USA \\
    $^2$University of Texas at Dallas, Richardson, TX, USA
}
\date{}

\begin{document}

\maketitle

\begin{abstract}
\label{sec:abstract}

Reinforcement learning (RL) in few-shot scenarios with limited sensor data is challenging due to insufficient training samples, particularly in applications like Dynamic Voltage and Frequency Scaling (DVFS) where sensor readings are semi-structured with inherent correlations. We propose Flow-Augmented Reinforcement Learning (FlowRL), a novel method that leverages continuous normalizing flows to generate high-quality synthetic data for few-shot RL. By integrating latent space bootstrapping for diversity and feature-weighted flow matching to preserve critical data correlations, FlowRL enhances sample efficiency and policy robustness. Evaluated on a DVFS case study using the NVIDIA Jetson TX2, our approach achieves up to 35\% higher frame rates and faster Q-value convergence compared to baselines, demonstrating its effectiveness in resource-constrained environments. FlowRL generalizes to other semi-structured domains, such as robotics and smart grids, offering a scalable solution for data-scarce RL settings.
\end{abstract}

\section{Introduction}
\label{sec:introduction}

The rapid advancements in large-scale generative models phrases have transformed data generation, particularly in unstructured domains like computer vision for generating images and videos. Techniques such as diffusion models \cite{ramesh2022hierarchical, saharia2022photorealistic, croitoru2023diffusion} have shown remarkable success in producing high-quality data, sparking interest in leveraging their underlying latent spaces for various applications. However, generating realistic and diverse semi-structured sensor data, which exhibits correlations and noise but lacks predefined relational structures, remains underexplored. This challenge is critical in few-shot reinforcement learning (RL), where models must generalize from limited examples. Addressing this is essential for advancing online RL in environments with semi-structured sensor data, such as in Dynamic Voltage and Frequency Scaling (DVFS), a power management technique that adjusts processor frequency and voltage to balance performance and energy efficiency using sensor readings like CPU frequency, temperature, power consumption, and frames per second (FPS). Complementary approaches include hierarchical multi-agent DVFS scheduling \cite{pivezhandi2026hidvfs}, zero-shot LLM-guided allocation \cite{pivezhandi2026zerodvfs}, statistical feature-aware task allocation \cite{pivezhandi2026feature}, and graph-driven performance modeling \cite{pivezhandi2026graphperf}.

In this paper, we introduce Flow-Augmented Reinforcement Learning (FlowRL), a novel approach designed to generate synthetic semi-structured sensor data tailored for few-shot RL. Unlike traditional data generation methods that require extensive datasets or rely on environment simulations, FlowRL creates diverse and realistic data samples from limited real-world trajectories. We achieve this through flow matching \cite{lipman2022flow}, a powerful method that enables simulation-free training of continuous normalizing flows (CNFs), improving the efficiency and quality of generated data compared to standard diffusion models. By integrating statistical techniques like bootstrapping \cite{efron1992bootstrap} into the flow matching process, we enhance latent space diversity, enabling better generalization across scenarios. Additionally, we use feature selection with Random Forests to prioritize critical data aspects, ensuring synthetic data captures essential correlations needed for effective RL training. While we draw inspiration from the Law of Large Numbers \cite{hastie2009elements} to guide distribution approximation, non-i.i.d. synthetic data limits strict convergence, so we rely on empirical validation to ensure practical improvements in policy performance.

FlowRL addresses the limitations of conventional model-based RL methods, which often struggle with overfitting and feature correlation when applied to semi-structured sensor data. We demonstrate that conventional methods increase feature correlation in synthetic data, reducing resemblance to real scenarios and limiting RL agent robustness in edge cases. By generating synthetic data that closely mirrors real-world sensor distributions, FlowRL enhances the adaptability and performance of RL models in few-shot settings. This is particularly valuable in dynamic, resource-constrained environments like DVFS, where real-time adaptability is critical. Through extensive experimentation on the NVIDIA Jetson TX2, we validate FlowRL, showing up to 35\% higher frame rates and faster Q-value convergence compared to baselines. FlowRL generalizes to other semi-structured domains, such as robotics and smart grids, offering a scalable solution for data-scarce RL applications.

\section{Preliminaries}
\label{sec:background}

This section introduces the core concepts underlying our approach to unstructured data augmentation for RL, leveraging flow matching to generate synthetic data that resembles unstructured data, with an emphasis on sample efficiency. We focus on unstructured data, such as sensor readings in our DVFS case study, which exhibit complex, non-grid-like patterns suitable for adapting methods designed for unstructured data. We cover Q-learning, regret bounds, the Law of Large Numbers (LLN), and the flow matching algorithm.

\subsection{Q-learning and Online RL}
Q-learning aims to learn a function $\mathcal{Q}(s,a)$ that estimates the expected \emph{return} (cumulative discounted reward) for taking action $a$ in state $s$ and following an optimal policy thereafter. For continuous states and discrete actions, a \textbf{Deep Q-Network (DQN)} approximates $\mathcal{Q}(s, a;\mathcal{W})$ with neural network parameters $\mathcal{W}$. The update rule is:
\(\mathcal{Q}(s_t, a_t; \mathcal{W}) \leftarrow \mathcal{Q}(s_t, a_t; \mathcal{W}) + \alpha \left[ r_t + \gamma \max_{a'} \mathcal{Q}(s_{t+1}, a'; \mathcal{W}^-) - \mathcal{Q}(s_t, a_t; \mathcal{W}) \right]\)
where $r_t$ is the immediate reward, $\gamma \in (0,1)$ is the discount factor, and $\mathcal{W}^-$ represents the parameters of a slower-moving \emph{target network}.

\paragraph{Online Setting.}
We focus on \textbf{online} RL, where the agent continuously interacts with the environment to refine its policy $\pi$. At each time-step, it observes $(s_t, a_t, r_t, s_{t+1})$ and updates $\mathcal{Q}$. \rev{In applications like DVFS, frequent decisions and gradual environmental changes allow treating each time-step as a horizon of $\mathcal{H}=1$, optimizing immediate returns for resource-management tasks.} This simplification is common in resource-management tasks where rewards are closely tied to instantaneous performance and power usage.

While Q-learning converges in tabular, finite-state settings under mild conditions, convergence in high-dimensional or partially observed environments is less assured. Using deep networks as function approximators can lead to instability or overfitting if sample diversity is limited.

\subsection{Regret Bounds and Sample Efficiency}
\label{subsec:regret-lln}
In RL, the \textbf{regret} after $T$ steps is defined as:
\[
\mathrm{Reg}(T) \;=\;
\sum_{t=1}^{T}
\Bigl[\,\mathcal{Q}^*(s, a) - \mathcal{Q}(s, a)\Bigr],
\]
representing the performance gap between the agent’s policy, using $\mathcal{Q}(s, a)$, and the optimal policy, using $\mathcal{Q}^*(s, a)$. Sublinear regret, such as $O(\sqrt{T})$, is achievable in \emph{finite-state}, well-explored environments \cite{jin2018q}, indicating that per-step regret diminishes over time.

\rev{In continuous, high-dimensional settings like our DVFS case study, achieving theoretical regret bounds is challenging.} If the function approximator or model class is misspecified (i.e., the true environment is outside the assumed class), sublinear bounds may not hold. In practice, we rely on empirical performance rather than strict theoretical guarantees.

\paragraph{Model Realizability and LLN.}
Sublinear regret often depends on \textbf{realizability}---the assumption that the true dynamics or $\mathcal{Q}^*$ function lies within the agent’s hypothesis space. In practice, limited exploration (e.g., rarely sampled CPU frequencies in DVFS) or complex environments can hinder the ability of neural networks to approximate the true dynamics, slowing or preventing convergence to near-optimal solutions. The \textit{Law of Large Numbers (LLN)} ensures that, with sufficiently \emph{diverse} and \emph{representative} data, empirical estimates (e.g., $\mathcal{Q}_T$ or transition model $P_T$) converge to their true values as $T \to \infty$. However, without effective exploration, convergence may remain incomplete.

\subsection{Flow Matching for Generative Modeling}
Flow matching \cite{lipman2022flow} is a generative method that models a continuous flow of probability density from a simple base distribution (e.g., $\mathcal{N}(0,I)$) at $t=0$ to a target distribution at $t=1$. The distribution evolves via:
\[
\frac{dx_t}{dt} \;=\; w(x_t, t),
\]
where $w(\cdot,\cdot)$ is a time-dependent velocity field for sample vector $x_t$ at time $t$. This field is approximated by a parametric network $v_{\mathcal{W}}$, trained to minimize:
\begin{equation}
\label{eq:fmeq_revised}
\min_{\mathcal{W}} \mathbb{E}_{x_t \sim p_t} \Bigl[\, \|v_{\mathcal{W}}(x_t, t) - w(x_t, t)\|^2 \Bigr].
\end{equation}

\paragraph{Relevance to DVFS RL.}
\rev{We employ flow matching to generate synthetic (state, action) tuples that capture the variability of unstructured data, such as sensor readings in our DVFS case study, using limited real data. By integrating Random Forest-based feature weighting, we ensure synthetic data closely resembles unstructured data, enhancing its suitability for RL training.} The convex training objective of flow matching improves sample efficiency, \rev{and our feature weighting approach produces robust synthetic data for RL when real-world samples are sparse.}

\section{Design of Distribution-Aware Flow Matching for Data Generation}
\label{sec:design}

In this section, we propose a \emph{distribution-aware} flow matching approach to generate synthetic unstructured data for online RL training with few shots, \rev{enhancing sample efficiency for applications like DVFS where data exhibits complex, semi-structured patterns.}

\subsection{Redefinition of the Regret Function}
To evaluate the impact of synthetic data in RL, we redefine the cumulative regret function to reflect the estimation error in the Bellman operator due to combined real and synthetic data, inspired by the model-free RL framework. Let $\mathcal{M}^*$ denote the true Markov Decision Process (MDP), with optimal Q-function $Q^*$. Let $\pi$ be the policy learned using real data $\mathcal{D}_R$ (size $n$) and synthetic data $\mathcal{D}_S$ (size $m$) from flow matching. The cumulative regret up to time $T$ is defined as:
\[
\mathrm{Reg}(T) = \sum_{t=1}^T \left[ Q^*(s_t, a_t) - Q^\pi(s_t, a_t) \right] + O( \sigma \sqrt{\frac{\log(1/\delta)}{n + m}}),
\]
where $Q^\pi$ is the Q-function under policy $\pi$, $\sigma^2$ bounds the variance of rewards and Q-values ($\mathrm{Var}[r + \gamma \max \mathcal{Q}] \leq \sigma^2$), $\delta$ is the confidence parameter, and the second term captures the estimation error of the empirical Bellman operator due to finite real and synthetic samples. This redefinition emphasizes the role of synthetic data in reducing variance in policy updates, consistent with the model-free regret bound.

\subsection{Flow Matching and Bootstrapping}
Our goal is to augment limited real trajectories, such as DVFS sensor data, with synthetic samples that approximate the unstructured data distribution, improving RL training efficiency. While synthetic samples may not be strictly i.i.d., flow matching empirically approximates real-world distributions, enhancing state-action coverage and policy performance.

Flow matching models the evolution of data sample probability density over a continuous-time parameter $t \in [0,1]$. \rev{Unlike model-based RL, which predicts next states $s_{t+1}$ given current states $s_t$ and actions $a_t$ using a learned environment model $\widehat{P}(s_{t+1} | s_t, a_t)$, flow matching generates synthetic tuples $(s_t, a_t, s_{t+1})$ directly from the joint distribution $p(s_t, a_t, s_{t+1})$. In its \emph{unconditional} form, the model learns a velocity field $v_{\mathcal{W}}(x_t)$ to match a target velocity field $w(x_t)$ \cite{tomczak2024deep}, where $x_t$ represents the tuple $(s_t, a_t, s_{t+1})$ at time $t$.} Equation \ref{eq:fmeq_revised} represents the baseline objective.

\paragraph{Bootstrapping in the Latent Space.}
To enhance diversity and mitigate overfitting, we introduce \emph{latent space bootstrapping}, inspired by \cite{efron1992bootstrap}. \rev{For a sample $x_0 \sim p_0(x)$ from the base latent distribution, representing an initial tuple $(s_0, a_0, s_1)$, we generate $B$ bootstrapped samples $\{ x_0^b \}_{b=1}^B$ via resampling:}
\[
x_0^b = \mathrm{Resample}(x_0), \quad b=1,\ldots,B.
\]
Let $x_t^b$ denote the sample evolved from $x_0^b$ at time $t$, with distribution $p_t^b(x)$. The bootstrapped flow matching objective extends Equation \ref{eq:fmeq_revised}:
\begin{equation}
\label{eq:fmbseq}
\min_{\mathcal{W}} \frac{1}{B} \sum_{b=1}^B \mathbb{E}_{\substack{t \in [0,1] \\ x_t^b \sim p_t^b(x)}} \Bigl[ \| v_{\mathcal{W}}(x_t^b) - w(x_t^b) \|^2 \Bigr].
\end{equation}
This objective promotes diverse synthetic trajectories by leveraging resampled latent distributions, ensuring broader coverage of the joint state-action-next-state space.

\subsection{Feature Weighting and Conditional Flow Matching}
Bootstrapping adds diversity, but \emph{feature weighting} ensures synthetic data resembles unstructured data for RL tasks, as required in our DVFS case study.

\paragraph{Random Forest Feature Importance.}
We employ a Random Forest regressor or classifier to assess how each real-world feature (e.g., sensor measurements in DVFS) contributes to predicting next-state transitions. Let $d$ be the dimension of our state/action/next-state representation. The feature importance at time $t$ for dimension (feature) $i$ is $\mathrm{Importance}(x_{t,i})$, which we normalize:
\begin{equation}
\label{eq:normweight}
\lambda_i = \frac{\mathrm{Importance}(x_{t,i})}{\sum_{j=1}^d \mathrm{Importance}(x_{t,j})}, \quad i = 1,\dots,d.
\end{equation}
In practice, the importance of each dimension (feature) is calculated based on how much the feature reduces impurity in classification or variance in regression tasks. The \textit{Importance} may be updated periodically or assumed to be roughly constant if the importance ranks are stable. Features with large $\lambda_i$ (e.g., temperature) are emphasized during training, preserving critical data relationships in synthetic samples.

\paragraph{Weighted Conditional Objective.}
We incorporate these weights into a \emph{conditional} flow matching setup, where additional context $z$ (e.g., rewards, auxiliary variables) follows $z \sim q(z)$. Define $x_{t,i}^b$ as the $i$-th feature of the bootstrapped sample at time $t$. The conditional flow matching objective with feature weighting extends \eqref{eq:fmbseq}:
\begin{equation}
\label{eq:cond_wbsfmeq}
\min_{\mathcal{W}} \frac{1}{B} \sum_{b=1}^B \mathbb{E}_{\substack{t \in [0,1]\\ x_t^b \sim p_t^b(x) \\ z \sim q(z)}} \Bigl[ \sum_{i=1}^d \lambda_i \| v_{\mathcal{W}}(x_{t,i}^b; z) - w(x_{t,i}^b; z) \|^2 \Bigr].
\end{equation}
\rev{This ensures high-importance features (e.g., temperature, resource usage) are prioritized, preserving essential correlations in the joint distribution $p(s_t, a_t, s_{t+1})$ for robust synthetic data generation in unstructured settings like DVFS.}

\subsection{Implications for RL: Regret Bounds under Model-Based Assumptions}
\label{subsec:rl_implications}

We theoretically motivate the use of synthetic data in a \textbf{model-based} RL framework. Let $\mathcal{M}^*$ be the true MDP that governs the DVFS transitions, and let $\widehat{\mathcal{M}}$ be a learned MDP model derived from real and synthetic data. \rev{In model-based RL, the environment model $\widehat{\mathcal{M}}$ predicts next states $s_{t+1}$ given states $s_t$ and actions $a_t$ via $\widehat{P}(s_{t+1} | s_t, a_t)$, unlike flow matching, which generates full tuples $(s_t, a_t, s_{t+1})$. Synthetic tuples from flow matching are used to train $\widehat{\mathcal{M}}$, enhancing its transition dynamics estimation.} The redefined regret function, $\mathrm{Reg}(T) = \sum_{t=1}^T [Q^*(s_t, a_t) - Q^\pi(s_t, a_t)] + O\left( \sigma \sqrt{\frac{\log(1/\delta)}{n + m}} \right)$, accounts for policy suboptimality and estimation error due to finite samples.

\paragraph{Enhanced Model Realizability.}
Synthetic data augments real data to improve the estimation of transition dynamics. The empirical Bellman operator for $\widehat{\mathcal{M}}$ is estimated using $\mathcal{D}_R$ and $\mathcal{D}_S$:
\[
\widehat{\mathcal{T}} \mathcal{Q}(s, a) = \frac{1}{n + m} ( \sum_{(s', r) \in \mathcal{D}_R} [r + \gamma \max_{a'} \mathcal{Q}(s', a')] 
\]
\[
+ \sum_{(s'', r) \in \mathcal{D}_S} [r + \gamma \max_{a'} \mathcal{Q}(s'', a')] ).
\]
Under bounded variance ($\mathrm{Var}[r + \gamma \max \mathcal{Q}] \leq \sigma^2$), a concentration bound gives, with probability $1 - \delta$:
\[
\| \widehat{\mathcal{T}} \mathcal{Q} - \mathcal{T} \mathcal{Q} \| \leq O\left( \sigma \sqrt{\frac{\log(1/\delta)}{n + m}} \right).
\]
This matches the regret second term, indicating that synthetic data reduces estimation error by increasing the effective sample size, enabling sublinear regret $\mathrm{Reg}(T) = O(\sqrt{T})$ under standard assumptions (finite horizon, bounded rewards) \cite{fan2021model}.

\paragraph{Illustrative Sketch.}
A simplified argument is:
\begin{enumerate}
    \item Each $(s,a)$ pair is sampled with probability $\mu > 0$ across real and synthetic data.
    \item The learned model $\widehat{\mathcal{M}}$ satisfies the above concentration bound.
    \item By standard analyses \cite{jin2018q,fan2021model}, regret satisfies:
    \[
    \mathrm{Reg}(T) = O(\sqrt{T}) + O\left( \sigma \sqrt{\frac{\log(1/\delta)}{n + m}} \right).
    \]
\end{enumerate}
The synthetic data contribution to $m$ ensures sublinear regret with reduced sample complexity.

\paragraph{Practical Caveats.}
In high-dimensional settings like DVFS, exploration is constrained, and real-synthetic mismatch may arise if the flow model fails to capture corner cases. Feature weighting mitigates this by ensuring synthetic tuples resemble unstructured data, aligning with the redefined regret.

\subsection{Implications for RL: Regret Bounds with Synthetic Data in Model-Free RL}
\label{subsec:rl_implications_modelfree}

In a model-free RL framework, synthetic data generated by flow matching augments real data to improve sample efficiency and reduce variance in Q-value estimates, leading to tighter regret bounds. \rev{Unlike model-based RL, which uses a learned model $\widehat{P}(s_{t+1} | s_t, a_t)$ to predict next states, flow matching directly generates synthetic tuples $(s_t, a_t, s_{t+1})$ from the joint distribution $p(s_t, a_t, s_{t+1})$, which are stored in $\mathcal{D}_S$ for Q-learning updates.}

\paragraph{Role of Synthetic Data via Flow Matching.}
Flow matching enhances the regret bound by generating synthetic tuples $(s_t, a_t, s_{t+1})$ that accurately capture the joint distribution $p(s_t, a_t, s_{t+1})$, expanding coverage of state-action-next-state combinations infrequently seen in real data. The bootstrapped objective (Equation \ref{eq:fmbseq}) ensures diversity, while the feature-weighted conditional objective (Equation \ref{eq:cond_wbsfmeq}) preserves critical correlations (e.g., temperature and frequency in DVFS). This reduces the variance term $O\left( \sigma \sqrt{\frac{\log(1/\delta)}{n + m}} \right)$ by increasing $m$, outperforming real-data-only training (where the term is $O\left( \sigma \sqrt{\frac{\log(1/\delta)}{n}} \right)$). By modeling continuous-time density evolution, flow matching minimizes distributional bias, ensuring synthetic tuples align with true dynamics, accelerating Q-value convergence and achieving tighter regret bounds.

Bias arises if synthetic tuples deviate from the true distribution. \rev{Feature weighting ensures fidelity to unstructured data distributions, keeping bias minimal, making flow matching-augmented model-free RL effective for few-shot settings, as validated in our DVFS experiments.}

\subsection{From LLN to Finite-Sample Guarantees}
\label{subsec:lln_finite}

Under mild mixing conditions, synthetic samples reduce the variance of empirical Bellman updates, yielding finite-sample bounds for both RL paradigms.

Let $\mathcal{D}_R$ be the $n$ real transitions and $\mathcal{D}_S$ the $m$ synthetic ones. The empirical backup is:
\[
\widehat{\mathcal{T}}\mathcal{Q}(s,a) = \frac{1}{n+m}\Bigl(\sum_{(s',r)\in \mathcal{D}_R} [r + \gamma\max_{a'}\mathcal{Q}(s',a')] 
\]
\[
+ \sum_{(s'',r)\in \mathcal{D}_S} [r + \gamma\max_{a'}\mathcal{Q}(s'',a')]\Bigr).
\]
With $\mathrm{Var}[r+\gamma\max \mathcal{Q}]\le \sigma^2$, a Bernstein-type bound gives, with probability $1-\delta$:
\[
|\widehat{\mathcal{T}}\mathcal{Q}(s,a) - \mathcal{T}\mathcal{Q}(s,a)| \leq O\left(\sigma\sqrt{\frac{\log(1/\delta)}{n+m}}\right) + O\left(\frac{\log(1/\delta)}{n+m}\right).
\]
Adding $m$ synthetic samples shrinks the variance term, accelerating convergence of Q-value or model estimates.

\subsection{Algorithmic Overview for Few-Shot Online RL}
\label{sec:alg_overview}

\rev{We integrate flow matching with a Deep Q-Network (DQN) using two replay buffers: $\mathcal{D}_{\mathrm{real}}$ for real tuples $(s_t, a_t, s_{t+1})$ collected from the environment, and $\mathcal{D}_{\mathrm{syn}}$ for synthetic tuples generated via flow matching.}

\begin{algorithm}[t]
\scriptsize
\begin{algorithmic}[1]
    \State \textbf{Initialize:} $\mathcal{D}_{\mathrm{real}}, \mathcal{D}_{\mathrm{syn}}$, DQN, $FM$
    \For{$i = 1$ \textbf{to} $T$}
        \State \rev{Collect real tuple $\tau_{\mathrm{real}} = (s_t, a_t, s_{t+1})$ and store in $\mathcal{D}_{\mathrm{real}}$}
        \If{enough new data in $\mathcal{D}_{\mathrm{real}}$}
            \State Train $FM$ using Equation~\eqref{eq:cond_wbsfmeq}
        \EndIf
        \For{$j=1$ \textbf{to rollout count}}
            \State \rev{$x_0^b = \mathrm{Resample}(x_0)$ \Comment{bootstrap tuple $(s_0, a_0, s_1)$}}
            \State \rev{Evolve $x_0^b$ to $x_t^b$ via $FM$ \Comment{generate tuple $(s_t, a_t, s_{t+1})$}}
            \State \rev{Store synthetic tuple $\tau_{\mathrm{syn}} = (s_t, a_t, s_{t+1})$ in $\mathcal{D}_{\mathrm{syn}}$}
        \EndFor
        \If{enough data in $\mathcal{D}_{\mathrm{real}}$ and $\mathcal{D}_{\mathrm{syn}}$}
            \State Sample mini-batches and update DQN
        \EndIf
    \EndFor
\end{algorithmic}
\caption{Distribution-Aware Flow Matching for Few-Shot Online RL.}
\label{alg:FlowMatchingDG}
\end{algorithm}

\rev{In each iteration, we collect real tuples $(s_t, a_t, s_{t+1})$ from the environment, storing them in $\mathcal{D}_{\mathrm{real}}$. When sufficient new data are available, we retrain or fine-tune the Flow-Matching model ($FM$) using the conditional weighted objective (Equation~\ref{eq:cond_wbsfmeq}) or the unconditional bootstrapped objective (Equation~\ref{eq:fmbseq}). We generate synthetic tuples by resampling bootstrapped latent points $\{x_0^b\}$, each representing an initial tuple $(s_0, a_0, s_1)$, evolving them over $t \in [0,1]$ via $FM$ to produce tuples $(s_t, a_t, s_{t+1})$, and storing these in $\mathcal{D}_{\mathrm{syn}}$. Mini-batches are sampled from both $\mathcal{D}_{\mathrm{real}}$ and $\mathcal{D}_{\mathrm{syn}}$ to update the DQN agent’s Q-function using the Q-learning update rule. This approach ensures robust state-action-next-state coverage for unstructured data, with bootstrapping enhancing diversity and feature weighting preserving critical correlations, supporting effective training for both model-based and model-free RL in few-shot settings, as validated in our DVFS case study.}

\section{Experiments}
\label{sec:experiments}
\begin{figure}[ht]
    \centering
    \includegraphics[width=.9\linewidth]{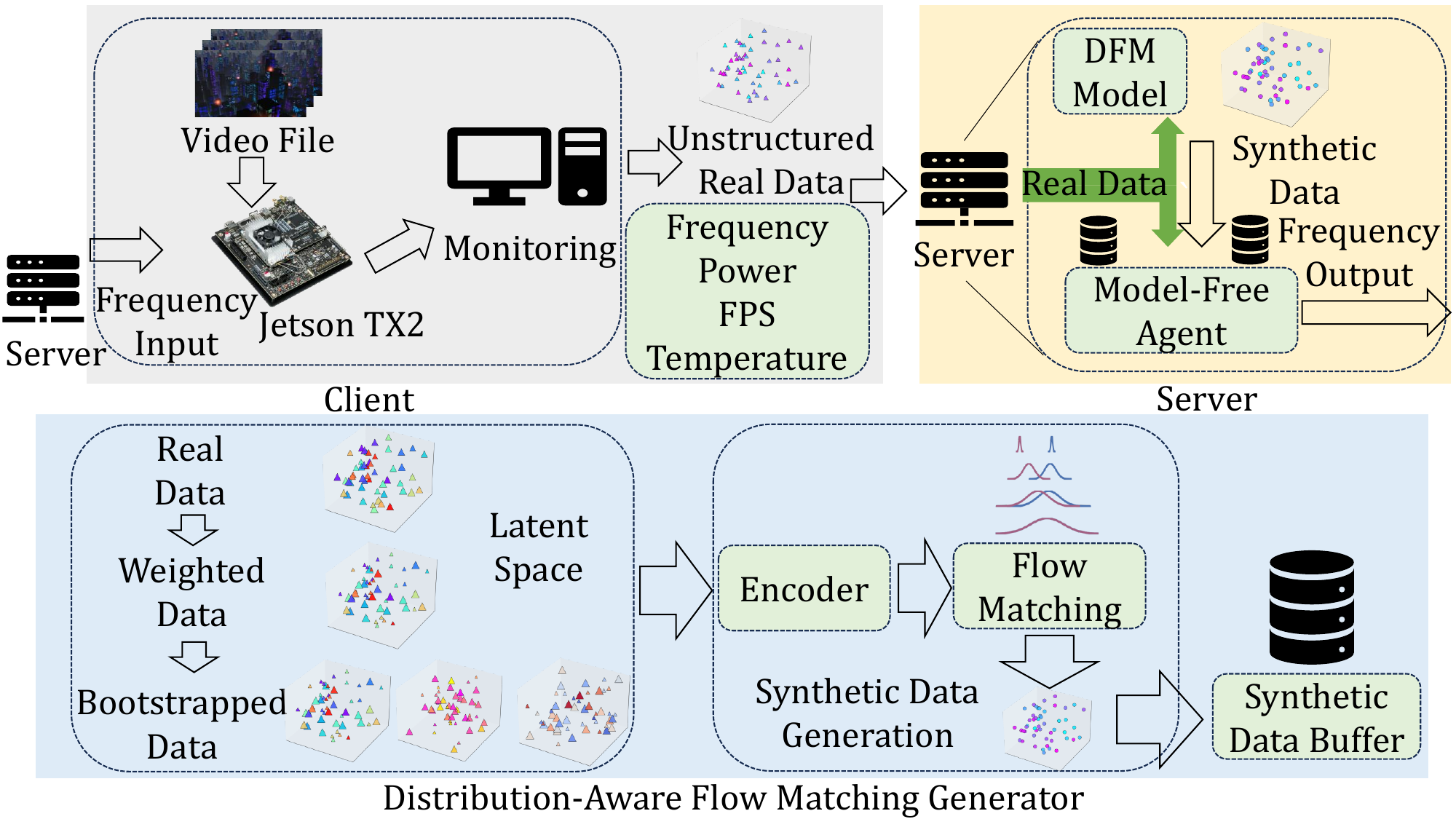}
    \caption{Workflow of the proposed Flow-Augmented Reinforcement Learning (FlowRL) approach.}
    \label{fig:workflow}
\end{figure}

In this section, we present the experimental setup, hyperparameter tuning, model architecture, and comparisons with model-based and model-free baselines to evaluate our Flow-Augmented Reinforcement Learning (FlowRL) approach.\paragraph{Experimental Setup}
We developed a Deep Q-Network (DQN) agent based on the zTT framework \cite{kim2021ztt} to optimize system performance under thermal and power constraints on the NVIDIA Jetson TX2, a power-efficient embedded platform with six heterogeneous CPU cores and an embedded GPU. The goal is to dynamically adjust frequency and power settings to maximize performance while ensuring system stability. The setup uses a client-server architecture: the client collects semi-structured state tuples \( \{FPS, f, \rho, \theta\} \) (frames per second, core frequency, power consumption, and core temperature), normalized via min-max scaling, and the server assigns frequency actions from a discrete set defined by hardware constraints (e.g., GPU\_CLOCK\_LIST). Each episode comprises 10 tuples collected at 1-second intervals over a 90-second horizon (90 decision points), with thermal monitoring triggering a 5–10-second sleep mode if temperature exceeds 50°C to mitigate throttling. The workflow is depicted in Figure~\ref{fig:workflow}.

\paragraph{Hyperparameter Tuning.}
We performed hyperparameter tuning using grid search, selecting values based on reward stability and convergence speed, as shown in Table \ref{tab:compact_hyperparameters}. The learning rate (0.05) was chosen after grid search and adjusted dynamically (e.g., reset to 1 under specific reward conditions), though it may seem high compared to typical deep learning ranges (e.g., 1e-4 to 5e-5). Future work will explore random search (e.g., Hyperopt) for refinement.

\begin{table}[h!]
    \centering
    \scriptsize
    \begin{tabular}{|l|c|}
        \hline
        \textbf{Hyperparameter} & \textbf{Tuning Values} \\ \hline
        \textbf{Experiment Time} & \{50, \textbf{90}, 200, 400, 800\} \\ \hline
        \textbf{Target FPS} & \{\textbf{30}, 60, 90\} \\ \hline
        \textbf{Target Temperature} & \{\textbf{50}, 60\} \\ \hline
        \textbf{Learning Rate} & \{0.01, \textbf{0.05}, 0.1\} \\ \hline
        \textbf{Discount Factor} & \{0.95, \textbf{0.99}\} \\ \hline
        \textbf{Epsilon Decay} & \{0.95, \textbf{0.99}\} \\ \hline
        \textbf{Batch Size} & \{\textbf{32}, 64, 128\} \\ \hline
        \textbf{Agent Train Start} & \{\textbf{32}, 40, 80, 100\} \\ \hline
        \textbf{Planning Iterations} & \{\textbf{100}, 200, 1000\} \\ \hline
        \textbf{Model Train Start} & \{\textbf{32}, 200, 300\} \\ \hline
        \textbf{Reward Scale ($\beta$)} & \{\textbf{0}, 2, 4\} \\ \hline
    \end{tabular}
    \caption{Main Hyperparameters and Tuning Values.}
    \label{tab:compact_hyperparameters}
\end{table}

\paragraph{Model Architecture and Training.}
The DQN agent features a neural network with two hidden layers of six neurons each, optimized using mean squared error (MSE) loss and the Adam optimizer. An $\epsilon$-greedy strategy starts with $\epsilon = 1.0$ and decays over time to promote exploitation. Training occurs in discrete intervals: Agent Train Start initiates Q-network updates, and Planning Iterations controls MDP model updates, using mini-batches of real and synthetic data. Performance metrics include Frames Per Second (FPS) for responsiveness, power consumption (\(\rho\)) for efficiency, average temperature (\(\theta\)) for safety, and average maximum Q-values for policy convergence, with emphasis on FPS and\(\beta\). The reward function, adapted from \cite{kim2021ztt}, is:

\[
R \;=\; u + v + \frac{\beta}{\rho},
\]
where
\[
u = \begin{cases} 
1 & \text{if } FPS \geq \text{target\_FPS}, \\
\frac{FPS}{\text{target\_FPS}} & \text{otherwise},
\end{cases}
\]

and

\[
v = \begin{cases} 
0.2 \cdot \tanh(\text{target\_temp} - \theta) & \text{if } \theta < \text{target\_temp}, \\ 
-2 & \text{if } \theta \geq \text{target\_temp}.
\end{cases}
\]

With target FPS at 30 and target temperature at 50°C, this design balances performance and constraints.

\paragraph{Model-Based and Model-Free Baselines.}
We compare FlowRL with DynaQ \cite{peng2018deep}, a model-based RL method; PlanGAN \cite{charlesworth2020plangan}, a generative model-based approach with dual-memory; MAML \cite{finn2017model}, a meta-learning method; and zTT \cite{kim2021ztt}, a model-free DQN baseline for DVFS. PlanGAN and DynaQ use real and synthetic data, while zTT runs directly on the Jetson TX2 without synthetic data. Consistent hyperparameters and reward functions ensure fair comparisons, with thermal throttling mitigated by sleep mode.

\paragraph{Results and Analysis.}
In our experiments, we train a \emph{conditional flow matching} model \cite{tomczak2024deep} on the real trajectories stored in a replay buffer, then use it to generate synthetic trajectories. We refer to this baseline generative approach as ``standard flow matching.'' We further refine it into a \emph{Flow-Augmented Reinforcement Learning (FlowRL)} method by incorporating latent-space bootstrapping and feature weighting to preserve important correlations and diversify the latent representation.We evaluate how well each synthetic approach captures correlations among features by computing Pearson's correlation coefficient \cite{sedgwick2012pearson}.  For two features $x_i$ and $x_j$ with $n$ samples, the coefficient is given by
\begin{equation}
\label{eq:pearson}
\text{corr}_{ij} = \frac{\sum_{k=1}^{n} (x_{i,k} - \bar{x}_i)\,(x_{j,k} - \bar{x}_j)}{\sqrt{\sum_{k=1}^{n} (x_{i,k} - \bar{x}_i)^2}\,\sqrt{\sum_{k=1}^{n} (x_{j,k} - \bar{x}_j)^2}}.
\end{equation}

\begin{figure}[ht]
    \centering
    \includegraphics[width=.65\linewidth]{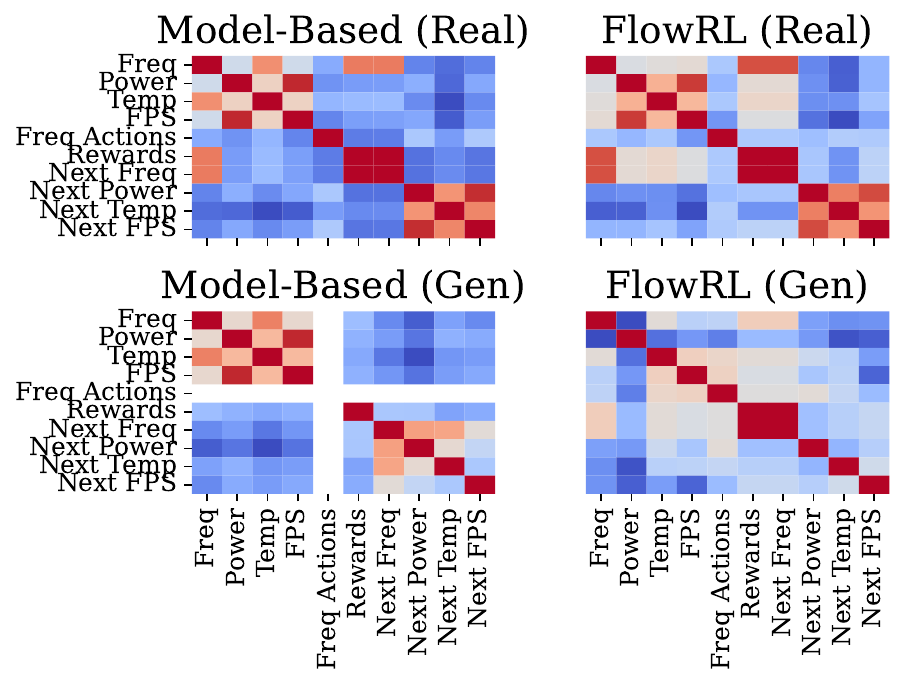}
    \caption{Correlation matrix comparisons.}
    \label{fig:correlation}
\end{figure}

When comparing the correlation matrices, we find that FlowRL-generated data better captures correlations among features visually compared to the model-based approach. Although the standard flow matching approach does capture some correlations, its matrix appears more uniform and less differentiated than that of FlowRL, suggesting that some critical feature relationships are weakened or diluted. Meanwhile, the model-based method shows stronger but sometimes misplaced correlations (dark regions in incorrect locations), indicating potential overfitting to specific patterns and misrepresenting genuine dependencies. FlowRL balances the two extremes by retaining essential correlations without artificially magnifying or suppressing them.

\begin{figure*}[ht]
    \centering
    \begin{minipage}[t]{0.55\textwidth}
        \centering
        \includegraphics[width=\linewidth]{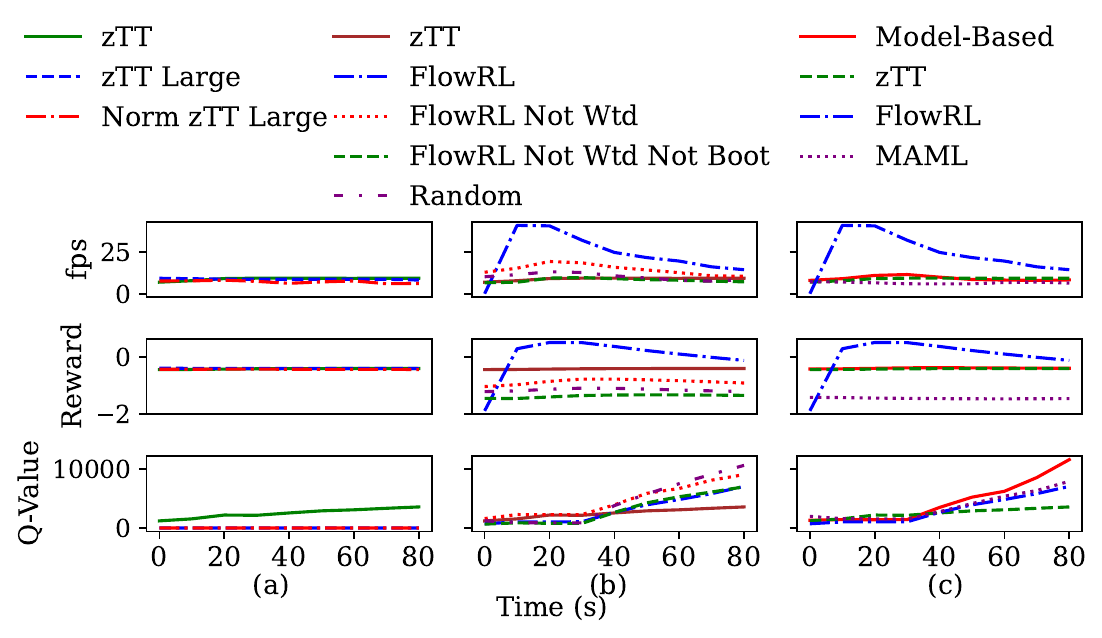}
        \caption{Performance comparisons across methods.}
        \label{fig:performance_comparison}
    \end{minipage}
    \hfill
    \begin{minipage}[t]{0.2\textwidth}
        \centering
        \includegraphics[width=\linewidth]{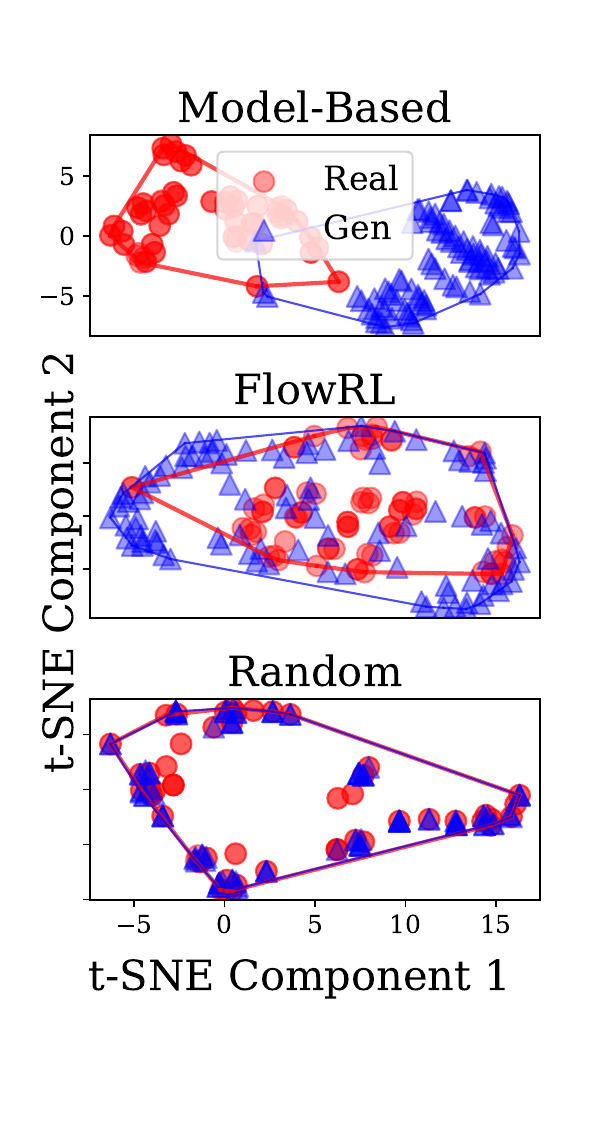}
        \caption{Real and (Gen)erated data t-SNE.}
        \label{fig:tsne_subplots_with_boundaries}
    \end{minipage}
\end{figure*}

To further analyze the data, we use t-SNE visualizations to compare the distributions in Figure~\ref{fig:tsne_subplots_with_boundaries}.

The t-SNE comparison of model-based and FlowRL with weighted and bootstrapped data and random data generation shows that FlowRL covers real data and more, while random sampling is restricted to the real data distribution and the model-based approach does not cover real data. This indicates FlowRL's superior ability to generalize beyond the original dataset. We also perform a Dynamic Time Warping (DTW) evaluation to assess temporal alignment. The DTW evaluation reveals that model-based data generation has a higher value (10,640) compared to FlowRL weighted and bootstrapped (7,341) between generated and real data, confirming FlowRL’s better temporal alignment.

To examine distributional coverage, we compare the histograms of key state-action features under different synthetic data generation methods.
\begin{figure}[ht]
    \centering
    \includegraphics[width=1\linewidth]{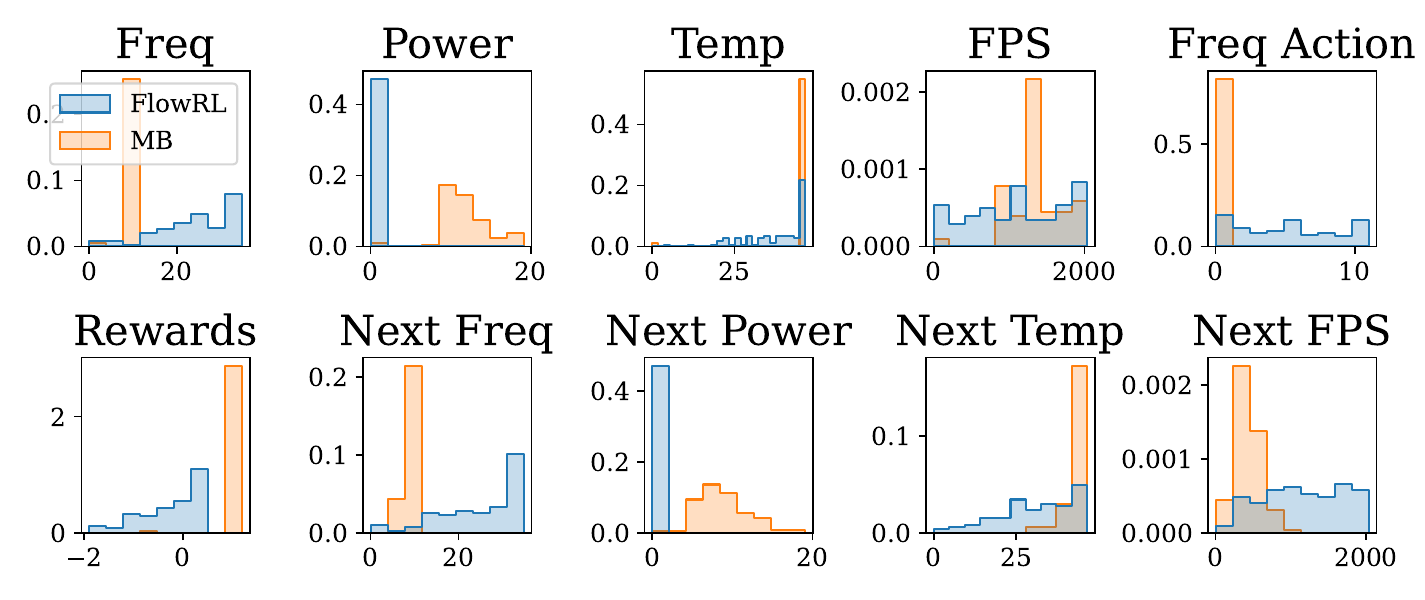}
    \caption{Comparisons of distributions of FlowRL and Model-Based (MB) generated data.}
    \label{fig:distribution}
\end{figure}
The results indicate that the FlowRL approach covers a wider spectrum of values compared to the model-based approach. In contrast, the model-based approach tends to be narrowly concentrated, particularly in power states and next states, suggesting it may overfit and fail to generalize to a broader range of real-world scenarios. The standard flow matching method similarly suffers from less coverage than FlowRL but does not exhibit the same extreme concentrations as the purely model-based approach. Overall, FlowRL’s expanded coverage helps the RL agent explore and learn policies that are more robust to varied conditions.

\paragraph{Performance Comparisons.}
To comprehensively evaluate FlowRL, three subplots were generated to compare performance metrics (average FPS, average reward, and average maximum Q-value) across different methods in 90 seconds, with the x-axis labeled in seconds (each second representing one iteration result based on FPS), as shown in Figure~\ref{fig:performance_comparison}. This figure includes a comprehensive ablation analysis of different implementations.

The left column of subplot (a) compares different model-free approaches based on zTT, comprising zTT, zTT with extended features from profiling data, and zTT with normalization, showing no specific differences in terms of average reward, Q-values, and frame rate, with stable FPS around 10. The middle figure (b) compares zTT with FlowRL, FlowRL without weighted but with bootstrapping (FM Not Wtd), FlowRL without augmentation, and random generated synthetic data augmented models, showing faster convergence in the FlowRL weighted bootstrapped version compared to all other approaches. The right column (c) extends the comparison to MAML, zTT, model-based, and FlowRL, where FlowRL outweighs all other approaches.

The above experiments confirm that incorporating bootstrapping and feature weighting into flow matching (i.e., FlowRL) improves both correlation preservation and distribution diversity in synthetic data generation. This yields more stable and high-performing RL policies compared to standard flow matching or purely model-based methods. The ability to balance performance and power highlights the potential of Flow-Augmented Reinforcement Learning for real-world DVFS tasks, where state-action spaces can be both large and only partially observed in limited data collection.
\section{Related Work}
\label{sec:relatedwork}
\paragraph{Machine Learning for DVFS.}
Machine learning has been widely adopted to optimize Dynamic Voltage and Frequency Scaling (DVFS) strategies, addressing the complexity of modern processors with adaptive control \cite{yu2020energy,kim2020autoscale,dinakarrao2019application,zhuo2021dvfs,pagani2018machine,shen2012learning,wang2017modular,yeganeh2020ring,liu2021cartad,sethi2021learning,ul2015hybrid,wang2021online,bo2021developing}. A survey by \cite{pagani2018machine} notes that model-free reinforcement learning (RL), such as the zTT framework \cite{kim2021ztt}, dominates, focusing on direct policy optimization for power-performance trade-offs. However, these methods often overlook feature correlation intricacies and the computational overhead of extensive data collection, particularly in data-scarce settings. Hierarchical multi-agent approaches \cite{pivezhandi2026hidvfs} and graph-driven performance models \cite{pivezhandi2026graphperf} have shown promise for DAG workloads. Our work builds on this by introducing FlowRL to enhance sample efficiency for semi-structured sensor data in DVFS.

\paragraph{Statistical Learning for DVFS.}
Statistical learning techniques have been explored to evaluate hardware events and application parameters impacting DVFS performance \cite{sasaki2007intra,cazorla2019probabilistic,liu2021cartad}. For example, \cite{sasaki2007intra} uses decision trees for energy-efficient DVFS lookup optimization, while \cite{cazorla2019probabilistic} assesses hardware counter importance for power reduction. \cite{liu2021cartad} applies f-score-based ranking to correlate compiler-level features with latency. These efforts, however, lack systematic consideration of runtime performance variability, sampling inefficiencies, and feature correlation effects on learning accuracy. FlowRL addresses these gaps by integrating feature weighting with Random Forests and data augmentation tailored for low-energy DVFS applications.

\paragraph{Bounding Sample Efficiency.}
Sample efficiency bounds for Q-learning have been a focus in RL research \cite{foster2023foundations,auer2008near,fan2021model,jin2018q}. Sublinear bounds have been established for finite state-action spaces \cite{auer2008near,jin2018q}, but extending these to continuous spaces requires robust function approximators. Approaches like \cite{fan2021model} rely on strong statistical assumptions, limiting real-world applicability. Deep Q-learning with stabilization techniques (e.g., experience replay, target networks) often faces instability \cite{foster2023foundations}, with few theoretical guarantees. Unlike these, FlowRL empirically validates sample efficiency for continuous semi-structured data in DVFS, offering practical insights over theoretical bounds.

\paragraph{Few-Shot RL.}
Few-shot RL methods, including transfer learning, meta-learning, and model-based approaches, aim to reduce data needs and enhance generalization \cite{wang2020generalizing,lee2020optimization,wang2016dueling,nachum2018data,florensa2017stochastic,choi2017multi,arora2021survey,wulfmeier2015deep,reddy2019sqil,vaswani2017attention}. Model-agnostic meta-learning (MAML) \cite{finn2017model} adapts to new tasks with minimal data, while model-based methods \cite{moerland2023model} approximate transition dynamics. DVFS-specific works like \cite{lin2023workload,kim2021ztt,zhou2021deadline,zhang2024dvfo} explore multi-agent RL and transfer learning but neglect statistical resampling or feature selection for energy optimization. Recent approaches address these gaps through statistical feature-aware task allocation \cite{pivezhandi2026feature} and zero-shot LLM-guided core allocation \cite{pivezhandi2026zerodvfs}. FlowRL advances this by using flow matching and bootstrapping to augment semi-structured sensor data, improving efficiency in DVFS and extending to robotics and smart grids.

\section{Conclusion}
\label{sec:conclusion}

We presented FlowRL, a novel method for few-shot RL with semi-structured sensor data. By leveraging flow matching, latent space bootstrapping, and feature-weighted conditioning, FlowRL generates high-quality synthetic data to enhance sample efficiency and policy robustness. Experiments on a DVFS task demonstrate up to 35\% higher frame rates and faster convergence compared to baselines like DQN, DDQN, and MAML. FlowRL’s applicability extends to other domains, such as robotics, offering a scalable solution for data-scarce RL. Future work includes exploring advanced RL algorithms (e.g., PPO) and neural-based feature weighting to further improve performance.

\bibliographystyle{unsrt}
\bibliography{aaai2026}

\begin{thebibliography}{10}

\bibitem{ramesh2022hierarchical}
Aditya Ramesh, Prafulla Dhariwal, Alex Nichol, Casey Chu, and Mark Chen.
\newblock Hierarchical text-conditional image generation with clip latents.
\newblock {\em arXiv preprint arXiv:2204.06125}, 1(2):3, 2022.

\bibitem{saharia2022photorealistic}
Chitwan Saharia, William Chan, Saurabh Saxena, Lala Li, Jay Whang, Emily~L
  Denton, Kamyar Ghasemipour, Raphael Gontijo~Lopes, Burcu Karagol~Ayan, Tim
  Salimans, et~al.
\newblock Photorealistic text-to-image diffusion models with deep language
  understanding.
\newblock {\em Advances in neural information processing systems},
  35:36479--36494, 2022.

\bibitem{croitoru2023diffusion}
Florinel-Alin Croitoru, Vlad Hondru, Radu~Tudor Ionescu, and Mubarak Shah.
\newblock Diffusion models in vision: A survey.
\newblock {\em IEEE Transactions on Pattern Analysis and Machine Intelligence},
  45(9):10850--10869, 2023.

\bibitem{pivezhandi2026hidvfs}
Mohammad Pivezhandi, Abusayeed Saifullah, and Ali Jannesari.
\newblock Hidvfs: A hierarchical multi-agent dvfs scheduler for openmp dag
  workloads.
\newblock {\em arXiv preprint}, 2026.

\bibitem{pivezhandi2026zerodvfs}
Mohammad Pivezhandi and Abusayeed Saifullah.
\newblock Zerodvfs: Zero-shot llm-guided core and frequency allocation for
  embedded platforms.
\newblock {\em arXiv preprint}, 2026.

\bibitem{pivezhandi2026feature}
Mohammad Pivezhandi, Abusayeed Saifullah, and Prashant Modekurthy.
\newblock Feature-aware task-to-core allocation in embedded multi-core
  platforms via statistical learning.
\newblock In {\em 2025 IEEE International Conference on Embedded and Real-Time
  Computing Systems and Applications (RTCSA)}, pages 102--113. IEEE, 2026.

\bibitem{pivezhandi2026graphperf}
Mohammad Pivezhandi, Mahdi Banisharif, Saeed Bakhshan, Abusayeed Saifullah, and
  Ali Jannesari.
\newblock Graphperf-rt: A graph-driven performance model for hardware-aware
  scheduling of openmp codes.
\newblock {\em arXiv preprint arXiv:2512.12091}, 2026.

\bibitem{lipman2022flow}
Yaron Lipman, Ricky~TQ Chen, Heli Ben-Hamu, Maximilian Nickel, and Matt Le.
\newblock Flow matching for generative modeling.
\newblock {\em arXiv preprint arXiv:2210.02747}, 2022.

\bibitem{efron1992bootstrap}
Bradley Efron.
\newblock Bootstrap methods: another look at the jackknife.
\newblock In {\em Breakthroughs in statistics: Methodology and distribution},
  pages 569--593. Springer, 1992.

\bibitem{hastie2009elements}
Trevor Hastie, Robert Tibshirani, Jerome~H Friedman, and Jerome~H Friedman.
\newblock {\em The elements of statistical learning: data mining, inference,
  and prediction}, volume~2.
\newblock Springer, 2009.

\bibitem{jin2018q}
Chi Jin, Zeyuan Allen-Zhu, Sebastien Bubeck, and Michael~I Jordan.
\newblock Is q-learning provably efficient?
\newblock {\em Advances in neural information processing systems}, 31, 2018.

\bibitem{tomczak2024deep}
Jakub~M Tomczak.
\newblock {\em Deep Generative Modeling}.
\newblock Springer Cham, 2024.

\bibitem{fan2021model}
Ying Fan and Yifei Ming.
\newblock Model-based reinforcement learning for continuous control with
  posterior sampling.
\newblock In {\em International Conference on Machine Learning}, pages
  3078--3087. PMLR, 2021.

\bibitem{kim2021ztt}
Seyeon Kim, Kyungmin Bin, Sangtae Ha, Kyunghan Lee, and Song Chong.
\newblock ztt: Learning-based dvfs with zero thermal throttling for mobile
  devices.
\newblock In {\em Proceedings of the 19th Annual International Conference on
  Mobile Systems, Applications, and Services}, pages 41--53, 2021.

\bibitem{peng2018deep}
Baolin Peng, Xiujun Li, Jianfeng Gao, Jingjing Liu, Kam-Fai Wong, and Shang-Yu
  Su.
\newblock Deep dyna-q: Integrating planning for task-completion dialogue policy
  learning.
\newblock {\em arXiv preprint arXiv:1801.06176}, 2018.

\bibitem{charlesworth2020plangan}
Henry Charlesworth and Giovanni Montana.
\newblock Plangan: Model-based planning with sparse rewards and multiple goals.
\newblock {\em Advances in Neural Information Processing Systems},
  33:8532--8542, 2020.

\bibitem{finn2017model}
Chelsea Finn, Pieter Abbeel, and Sergey Levine.
\newblock Model-agnostic meta-learning for fast adaptation of deep networks.
\newblock In {\em International conference on machine learning}, pages
  1126--1135. PMLR, 2017.

\bibitem{sedgwick2012pearson}
Philip Sedgwick.
\newblock Pearson’s correlation coefficient.
\newblock {\em Bmj}, 345, 2012.

\bibitem{yu2020energy}
Zheqi Yu, Pedro Machado, Adnan Zahid, Amir~M Abdulghani, Kia Dashtipour, Hadi
  Heidari, Muhammad~A Imran, and Qammer~H Abbasi.
\newblock Energy and performance trade-off optimization in heterogeneous
  computing via reinforcement learning.
\newblock {\em Electronics}, 9(11):1812, 2020.

\bibitem{kim2020autoscale}
Young~Geun Kim and Carole-Jean Wu.
\newblock Autoscale: Energy efficiency optimization for stochastic edge
  inference using reinforcement learning.
\newblock In {\em 2020 53rd Annual IEEE/ACM international symposium on
  microarchitecture (MICRO)}, pages 1082--1096. IEEE, 2020.

\bibitem{dinakarrao2019application}
Sai Manoj~Pudukotai Dinakarrao, Arun Joseph, Anand Haridass, Muhammad Shafique,
  J{\"o}rg Henkel, and Houman Homayoun.
\newblock Application and thermal-reliability-aware reinforcement learning
  based multi-core power management.
\newblock {\em ACM Journal on Emerging Technologies in Computing Systems
  (JETC)}, 15(4):1--19, 2019.

\bibitem{zhuo2021dvfs}
Cheng Zhuo, Di~Gao, Yuan Cao, Tianhao Shen, Li~Zhang, Jinfang Zhou, and Xunzhao
  Yin.
\newblock A dvfs design and simulation framework using machine learning models.
\newblock {\em IEEE Design \& Test}, 2021.

\bibitem{pagani2018machine}
Santiago Pagani, PD~Sai Manoj, Axel Jantsch, and J{\"o}rg Henkel.
\newblock Machine learning for power, energy, and thermal management on
  multicore processors: A survey.
\newblock {\em IEEE Transactions on Computer-Aided Design of Integrated
  Circuits and Systems}, 39(1):101--116, 2018.

\bibitem{shen2012learning}
Hao Shen, Jun Lu, and Qinru Qiu.
\newblock Learning based dvfs for simultaneous temperature, performance and
  energy management.
\newblock In {\em Thirteenth International Symposium on Quality Electronic
  Design (ISQED)}, pages 747--754. IEEE, 2012.

\bibitem{wang2017modular}
Zhe Wang, Zhongyuan Tian, Jiang Xu, Rafael~KV Maeda, Haoran Li, Peng Yang,
  Zhehui Wang, Luan~HK Duong, Zhifei Wang, and Xuanqi Chen.
\newblock Modular reinforcement learning for self-adaptive energy efficiency
  optimization in multicore system.
\newblock In {\em 2017 22nd Asia and South Pacific Design Automation Conference
  (ASP-DAC)}, pages 684--689. IEEE, 2017.

\bibitem{yeganeh2020ring}
Amir Yeganeh-Khaksar, Mohsen Ansari, Sepideh Safari, Sina Yari-Karin, and
  Alireza Ejlali.
\newblock Ring-dvfs: Reliability-aware reinforcement learning-based dvfs for
  real-time embedded systems.
\newblock {\em IEEE Embedded Systems Letters}, 13(3):146--149, 2020.

\bibitem{liu2021cartad}
Di~Liu, Shi-Gui Yang, Zhenli He, Mingxiong Zhao, and Weichen Liu.
\newblock Cartad: Compiler-assisted reinforcement learning for thermal-aware
  task scheduling and dvfs on multicores.
\newblock {\em IEEE Transactions on Computer-Aided Design of Integrated
  Circuits and Systems}, 2021.

\bibitem{sethi2021learning}
Udhav Sethi.
\newblock Learning energy-aware transaction scheduling in database systems.
\newblock Master's thesis, University of Waterloo, 2021.

\bibitem{ul2015hybrid}
Fakhruddin Muhammad~Mahbub ul~Islam and Man Lin.
\newblock Hybrid dvfs scheduling for real-time systems based on reinforcement
  learning.
\newblock {\em IEEE Systems Journal}, 11(2):931--940, 2015.

\bibitem{wang2021online}
Yiming Wang, Weizhe Zhang, Meng Hao, and Zheng Wang.
\newblock Online power management for multi-cores: A reinforcement learning
  based approach.
\newblock {\em IEEE Transactions on Parallel and Distributed Systems},
  33(4):751--764, 2021.

\bibitem{bo2021developing}
Zitong Bo, Ying Qiao, Chang Leng, Hongan Wang, Chaoping Guo, and Shaohui Zhang.
\newblock Developing real-time scheduling policy by deep reinforcement
  learning.
\newblock In {\em 2021 IEEE 27th Real-Time and Embedded Technology and
  Applications Symposium (RTAS)}, pages 131--142. IEEE, 2021.

\bibitem{sasaki2007intra}
Hiroshi Sasaki, Yoshimichi Ikeda, Masaaki Kondo, and Hiroshi Nakamura.
\newblock An intra-task dvfs technique based on statistical analysis of
  hardware events.
\newblock In {\em Proceedings of the 4th international conference on Computing
  frontiers}, pages 123--130, 2007.

\bibitem{cazorla2019probabilistic}
Francisco~J Cazorla, Leonidas Kosmidis, Enrico Mezzetti, Carles Hernandez,
  Jaume Abella, and Tullio Vardanega.
\newblock Probabilistic worst-case timing analysis: Taxonomy and comprehensive
  survey.
\newblock {\em ACM Computing Surveys (CSUR)}, 52(1):1--35, 2019.

\bibitem{foster2023foundations}
Dylan~J Foster and Alexander Rakhlin.
\newblock Foundations of reinforcement learning and interactive decision
  making.
\newblock {\em arXiv preprint arXiv:2312.16730}, 2023.

\bibitem{auer2008near}
Peter Auer, Thomas Jaksch, and Ronald Ortner.
\newblock Near-optimal regret bounds for reinforcement learning.
\newblock {\em Advances in neural information processing systems}, 21, 2008.

\bibitem{wang2020generalizing}
Yaqing Wang, Quanming Yao, James~T Kwok, and Lionel~M Ni.
\newblock Generalizing from a few examples: A survey on few-shot learning.
\newblock {\em ACM computing surveys (csur)}, 53(3):1--34, 2020.

\bibitem{lee2020optimization}
Donghwan Lee, Niao He, Parameswaran Kamalaruban, and Volkan Cevher.
\newblock Optimization for reinforcement learning: From a single agent to
  cooperative agents.
\newblock {\em IEEE Signal Processing Magazine}, 37(3):123--135, 2020.

\bibitem{wang2016dueling}
Ziyu Wang, Tom Schaul, Matteo Hessel, Hado Hasselt, Marc Lanctot, and Nando
  Freitas.
\newblock Dueling network architectures for deep reinforcement learning.
\newblock In {\em International conference on machine learning}, pages
  1995--2003. PMLR, 2016.

\bibitem{nachum2018data}
Ofir Nachum, Shixiang~Shane Gu, Honglak Lee, and Sergey Levine.
\newblock Data-efficient hierarchical reinforcement learning.
\newblock {\em Advances in neural information processing systems}, 31, 2018.

\bibitem{florensa2017stochastic}
Carlos Florensa, Yan Duan, and Pieter Abbeel.
\newblock Stochastic neural networks for hierarchical reinforcement learning.
\newblock {\em arXiv preprint arXiv:1704.03012}, 2017.

\bibitem{choi2017multi}
Jinyoung Choi, Beom-Jin Lee, and Byoung-Tak Zhang.
\newblock Multi-focus attention network for efficient deep reinforcement
  learning.
\newblock {\em arXiv preprint arXiv:1712.04603}, 2017.

\bibitem{arora2021survey}
Saurabh Arora and Prashant Doshi.
\newblock A survey of inverse reinforcement learning: Challenges, methods and
  progress.
\newblock {\em Artificial Intelligence}, 297:103500, 2021.

\bibitem{wulfmeier2015deep}
Markus Wulfmeier, Peter Ondruska, and Ingmar Posner.
\newblock Deep inverse reinforcement learning.
\newblock {\em CoRR, abs/1507.04888}, 2015.

\bibitem{reddy2019sqil}
Siddharth Reddy, Anca~D Dragan, and Sergey Levine.
\newblock Sqil: Imitation learning via reinforcement learning with sparse
  rewards.
\newblock {\em arXiv preprint arXiv:1905.11108}, 2019.

\bibitem{vaswani2017attention}
Ashish Vaswani, Noam Shazeer, Niki Parmar, Jakob Uszkoreit, Llion Jones,
  Aidan~N Gomez, {\L}ukasz Kaiser, and Illia Polosukhin.
\newblock Attention is all you need.
\newblock {\em Advances in neural information processing systems}, 30, 2017.

\bibitem{moerland2023model}
Thomas~M Moerland, Joost Broekens, Aske Plaat, Catholijn~M Jonker, et~al.
\newblock Model-based reinforcement learning: A survey.
\newblock {\em Foundations and Trends{\textregistered} in Machine Learning},
  16(1):1--118, 2023.

\bibitem{lin2023workload}
Chengdong Lin, Kun Wang, Zhenjiang Li, and Yu~Pu.
\newblock A workload-aware dvfs robust to concurrent tasks for mobile devices.
\newblock In {\em Proceedings of the 29th Annual International Conference on
  Mobile Computing and Networking}, pages 1--16, 2023.

\bibitem{zhou2021deadline}
Ti~Zhou and Man Lin.
\newblock Deadline-aware deep-recurrent-q-network governor for smart energy
  saving.
\newblock {\em IEEE Transactions on Network Science and Engineering},
  9(6):3886--3895, 2021.

\bibitem{zhang2024dvfo}
Ziyang Zhang, Yang Zhao, Huan Li, Changyao Lin, and Jie Liu.
\newblock Dvfo: Learning-based dvfs for energy-efficient edge-cloud
  collaborative inference.
\newblock {\em IEEE Transactions on Mobile Computing}, 2024.

\end{thebibliography}

\end{document}